\documentclass{INTERSPEECH2023}
\usepackage{multirow}
\usepackage{hyperref}

\interspeechcameraready

\title{NoRefER: a Referenceless Quality Metric for Automatic Speech Recognition via Semi-Supervised Language Model Fine-Tuning with Contrastive Learning}
\name{Kamer Ali Yuksel, 
      Thiago Ferreira, 
      Golara Javadi, 
      Mohamed El-Badrashiny, 
      Ahmet Gunduz
      }
\address{aiXplain Inc., Los Gatos, CA, USA}
\email{\{kamer, thiago, golara, Mohamed, ahmet\}@aixplain.com}

\begin{document}

\maketitle
 
\begin{abstract}
This paper introduces NoRefER, a novel referenceless quality metric for automatic speech recognition (ASR) systems. Traditional reference-based metrics for evaluating ASR systems require costly ground-truth transcripts. NoRefER overcomes this limitation by fine-tuning a multilingual language model for pair-wise ranking ASR hypotheses using contrastive learning with Siamese network architecture. The self-supervised NoRefER exploits the known quality relationships between hypotheses from multiple compression levels of an ASR for learning to rank intra-sample hypotheses by quality, which is essential for model comparisons. The semi-supervised version also uses a referenced dataset to improve its inter-sample quality ranking, which is crucial for selecting potentially erroneous samples. The results indicate that NoRefER correlates highly with reference-based metrics and their intra-sample ranks, indicating a high potential for referenceless ASR evaluation or a/b testing.
\end{abstract}

\section{Introduction}
\label{sec:intro}
Automatic Speech Recognition (ASR) is considered to be the future means of communication between humans and machines. This field has made significant progress over the last few years with the advent of Deep Neural Networks~\cite{malik2021automatic, aldarmaki2022unsupervised}. Modern ASR systems rely on large amounts of annotated speech to learn accurate speech representation and recognition, and they can achieve remarkable accuracy for resource-rich languages like English. The performance of ASR systems usually depends on two factors, the accuracy of the output produced as well as the processing speed of the ASR. The most commonly used evaluation metric for ASR systems is Word-Error-Rate (WER), which calculates errors on the word level and requires annotated resources as ground-truth transcripts that may not always be available or accurately reflect the ASR output's quality.

\begin{figure*}[t]
\centering 
\centerline{\includegraphics[width=\textwidth]{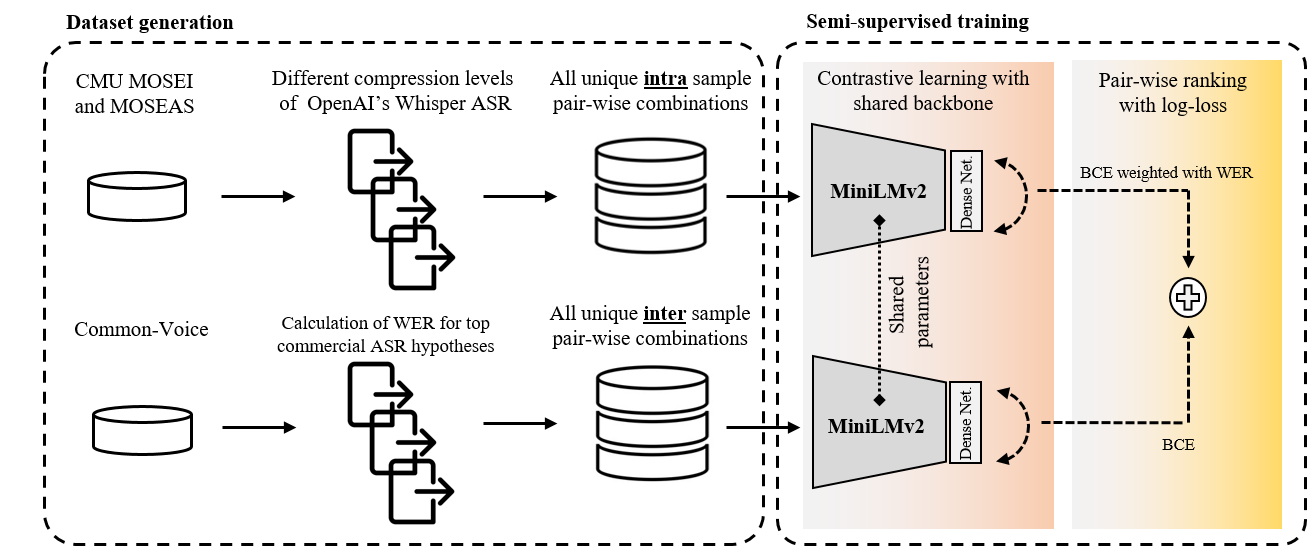}}
\caption{General framework of NoRefER. i) Dataset generation. Six compression levels of OpenAI's Whisper~\cite{radford2022robust} are used as ASR models to compute all pair-wise combinations of different quality outcomes. ii) Multi-task supervised-learning with self-supervised learning in pairwise ranking. It fine-tunes a pre-trained language model with contrastive learning of semi-supervised hypothesis pairs.}
\label{fig:frame}
\end{figure*}

ASR quality estimation without transcripts follows a two-stage framework, including feature extraction and WER prediction. Most of the research in this field used hand-crafted features to build linear regression-based algorithms~\cite{fan2019neural}. NoRefER, a novel referenceless quality metric for ASR systems, is introduced in this paper to reduce the necessity of manual processes, and to present a measure for evaluating ASR performance with limited or no ground-truth references. The NoRefER referenceless quality metric provides a much-needed solution to the limitations of traditional reference-based metrics through fine-tuning a multi-language language model (LM) with contrastive learning and pair-wise ranking (Fig.~\ref{fig:frame}). First, a referenceless training dataset of ASR hypothesis pairs is formed from the pairwise combinations of unique outputs of OpenAI’s Whisper ASR model~\cite{radford2022robust} with six different compression levels. The higher the compression level gets, the lower the quality is expected. This dataset is used for self-supervised contrastive learning using a Siamese architecture~\cite{Chen_2021_CVPR}. Then, The fine-tuned LM is used as a shared backbone with the supervised task on a referenced dataset. The two pairwise ranking branches are used to improve the performance of the proposed metric for inter and intra-sample ranking by leveraging both datasets. The intra-sample and inter-sample pair-wise quality ranking decisions of the referenceless metric are validated on several blind test datasets in various languages in comparison with the perplexity metric from XLM-RoBERTa-Large~\cite{conneau2019unsupervised}. All models and source-codes required for reproducing the experimental results are available here: \url{https://anonymous.4open.science/r/Interspeech-NoRefER-6887}

The contributions from this paper can be listed as the followings: \textbf{i)} a novel method for self-supervision of the ASR hypothesis quality by utilizing multiple compression levels of an ASR model; \textbf{ii)} a multi-language referenceless ASR quality estimation metric through an LM fine-tuned with contrastive learning using a Siamese network architecture and pair-wise ranking; \textbf{iii)} applying semi-supervised learning to improve the self-supervised metric's performance on inter-sample ranking by using two corpora with and without reference transcriptions; \textbf{iv)} demonstrating the significant potential of the proposed metric against the perplexity from the state-of-art multi-lingual LM, regarding their correlation with WER scores and ranks.

\section{Related Work}

Reference-based evaluation metrics are among the most widely used methods to evaluate the performance of ASR systems. The most common metrics are WER and phoneme error rate (PER). WER measures the proportion of words in the reference that are not recognized correctly, whereas PER measures the proportion of phonemes that are not recognized correctly~\cite{rabiner1993fundamentals}. Both metrics require a ground-truth reference transcript to compare the ASR(s) outputs. Their main limitation is the dependence on the existence of a reference transcript, where a poor reference can also negatively affect the accuracy of the evaluation. 

Referenceless evaluation metrics for ASR do not rely on a ground-truth reference. Some existing referenceless evaluation metrics include confidence scores and confidence-based fusion~\cite{kalgaonkar2015estimating,swarup2019improving, qiu2021learning}. Confidence scores are usually generated by the ASR system and are used to rank alternative hypotheses. These scores are not practical for the black-box ASR systems unless the model is designed to output the confidence score. The confidence-based fusion combines multiple outputs from different ASR systems to produce a single output with improved quality. However, they are dependent on the particular ASR system being used and cannot capture the full context of the ASR output; and therefore, they may not be accurate in some instances.

 Previous research on ASR quality estimation has mainly focused on supervised regression or ordinal classification of speech and language features~\cite{del2018speaker,swarup2019improving,jiang2005confidence,kalgaonkar2015estimating}. For example, \cite{fan2019neural} used a bidirectional transformer language model to estimate ASR quality by modeling the empirical distribution of WER through a neural zero-inflated Beta regression layer. Ali and Renals~\cite{ali2020word} who did not have access to the ASR system, utilized a multistream end-to-end architecture with acoustic, lexical, and phonotactic features to estimate WER. Sheshadri \textit{et al.}~\cite{sheshadri2021bert} proposed a BERT-based architecture with speech features for balanced ordinal classification to estimate WER. However, none of these referenceless ASR quality estimators were solely based on language features and were not trained with references. On the other hand, Namazifar \textit{et al.}~\cite{namazifar2021correcting} leveraged the robustness of warped language models against transcription noise to correct the transcriptions of spoken language, which resulted in a 10\% reduction in WER for both automatic and manual transcriptions. While they could use the distance with improved transcription for ASR quality estimation, they did not study that use-case.

 Lastly, the WMT Quality Estimation Shared Task \cite{mathur2020results} is a well-known evaluation framework for quality estimation (QE) metrics in machine translation (MT). The task recently also includes ranking the quality of machine-generated translations without access to reference translations. This is done by training quality estimation models on parallel sentences with human-annotated quality scores. As an outcome of the WMT Shared Task, various referenceless QE metrics have emerged in the MT domain, including COMET-QE~\cite{rei-etal-2020-unbabels}. COMET-QE also utilizes contrastive-learning to fine-tune a pre-trained multi-lingual LM for MT quality estimation to distinguish between high and low-quality parallel MT hypotheses. However, the fine-tuning of COMET-QE relies on and is limited by the existence of a human-evaluation dataset or ground-truth references. Whereas, in this work, self-supervision in contrastive-learning is achieved via exploiting the known quality relationships without costly human annotations to compose a training dataset for the metric.

\section{Methodology}
\label{sec:methodology}
The proposed method utilizes a pre-trained language model that is fine-tuned using contrastive learning, employing a Siamese network architecture for pair-wise quality ranking decisions. To achieve this, multi-task learning is utilized, with a shared backbone that processes unique pair combinations from two sets of datasets for training and validation. The first task focuses on self-supervised learning by exploiting known quality relationships between multiple compression levels of an ASR model. The second task employs supervised learning for inter-sample comparison, utilizing a dataset with ground-truth. By multi-tasking self-supervised and supervised learning, the proposed method offers a semi-supervised approach to train a LM for the quality ranking of the intra- and inter-sample hypotheses.

\subsection{Dataset generation}
\label{sec:datasetprep}
To train and validate the proposed referenceless quality metric with self-supervision, unique outputs from an ASR model~\cite{radford2022robust} are utilized to form pairwise combinations that can be used for contrastive-learning. Multiple compression levels are used as a proxy for quality, where higher model compression levels indicates lower-quality transcriptions. The process of extracting unique pairs involves selecting two ASR hypotheses, one with higher quality and one with lower quality, for the same speech and combining them into a single pair. These pairs are shuffled and placed into mini-batches for training and validation sets. Inconsistent pairs, for which the exact reverse pair also exists, are dropped. Each pair's training and validation loss is weighted using the Word Error Rate (WER) between the paired hypotheses. The model is penalized more for incorrect pair-wise ranking decisions when the distance between two hypotheses is high, as it is more acceptable to make a pair-wise ranking mistake when they are close. By fine-tuning the proposed Siamese network using this approach, the quality metric can be trained and validated effectively without requiring ground-truth transcriptions.

\subsection{Self-supervised learning}
\label{sec:self}
The proposed method utilizes a pre-trained cross-lingual language model (LM) with a Siamese network architecture, followed by a dense encoder that reduces the embeddings produced by the LM to a single scalar logit. This logit is then used to compare the outputs of the Siamese network. The pre-trained LM in this architecture is MiniLMv2, a smaller and (2.7x) faster language understanding model with 117M parameters distilled from XLM-RoBERTa-Large~\cite{wang2020minilmv2}, which has 560M parameters. The dense encoder has two linear layers with a 10\% dropout ratio and a non-linear activation function in-between.

To fine-tune the pre-trained LM on a pair-wise ranking task with contrastive learning, a self-supervised learning method that trains a model to distinguish between positive and negative examples, the proposed method utilizes ASR outputs to generate pairs that can be compared to predict the one with higher quality. The contrastive-learning process uses the shared network to take a pair as input and output a logit for each. These logits are then subtracted from each other, and a Sigmoid activation function is applied to their difference to produce a probability for binary classification of their qualities. The Adafactor~\cite{shazeer2018adafactor} optimizer is used with its default parameters and a learning rate of 1e-5 for fine-tuning the LM on this pair-wise ranking task using Binary Cross-Entropy (Log-Loss) weighted by the WER in-between pairs. Eq.~\ref{eq:selfloss} shows the self-supervised loss.

\begin{align}
  \mathcal{L}_{self} = -\frac{1}{N}\sum_{i=1}^{N}\mathcal{W}(P_{i})\log(\sigma(\mathcal{M}(P_{i_{+}})-\mathcal{M}(P_{i_{-}}))) 
  \label{eq:selfloss}
\end{align}
where $\mathcal{W}(.)$ is the WER in-between pairs, $\sigma(.)$ is the Sigmoid activation, $\mathcal{M}(.)$ is the output of the model. N is the number of pairs, and $P_{i_{+}}$ and $P_{i_{-}}$ show the positive and negative samples of each pair, respectively. By utilizing this contrastive-learning process, the LM can learn a high-level representation of pairs that is discriminative of their quality. This approach can effectively train a referenceless quality metric, allowing accurate intra-sample hypothesis comparisons without references.

\subsection{Semi-supervised learning}
\label{sec:semi}
The NoRefER method extends its self-supervised fine-tuning to a semi-supervised version by introducing an additional Log-Loss term (Fig.~\ref{fig:frame}). The supervised batch of pairs used for this task includes pairwise combinations of inter-sample hypotheses for non-parallel speech samples, where the quality relationship and the reference are known and can be quantified using reference-based ASR quality metrics (such as WER). In contrast to the self-supervised loss term, the losses of the supervised pairs are not weighted by the distance between hypotheses but by the positive-class weight of randomly formed pairs. To maximize the efficiency of the limited supervised hypotheses, the pairwise combinations of inter-sample hypotheses are formed by self-concatenating each mini-batch of supervised hypotheses, shuffling the concatenated column, and re-assigning the pairwise ranking binary classification labels based on the WERs calculated from ground-truth references. This approach randomizes the formation of pairwise combinations of supervised inter-sample hypotheses during the training process. With this framework, the training loss is comprised of two objectives: the self-supervised loss ($\mathcal{L}_{self}$) on a generated referenceless dataset, and the supervised loss ($\mathcal{L}_{sup}$) on supervised hypotheses. Eq.~\ref{eq:semiloss} shows the supervised loss, and Eq.~\ref{eq:totloss} shows the sum of both losses for multi-tasking in semi-supervised learning.

\begin{align}
  \mathcal{L}_{sup} = -\frac{1}{N}\sum_{i=1}^{N}\log(\sigma(\mathcal{M}(P_{i_{+}})-\mathcal{M}(P_{i_{-}})))
  \label{eq:semiloss}
\end{align}
\begin{align}
  \mathcal{L}_{SEMI} = \alpha \mathcal{L}_{sup} + (1-\alpha)\mathcal{L}_{self}
  \label{eq:totloss}
\end{align}
where $\alpha$ is a weight parameter for combining the two losses and was set to 0.5 by default. The supervised and self-supervised mini-batches are equal and set to 128 during the training. This semi-supervised training enables NoRefER to learn from a large amount of self-supervised and a limited amount of supervised information, leading to improved performance on the referenceless ASR quality estimation. Combining the self-supervised and supervised information enables the model to generalize better to unseen data, resulting in a more robust and accurate referenceless ASR quality metric. This is particularly useful for comparing non-parallel hypotheses, e.g. when prioritizing production hypotheses for post-editing in active-learning while extending the ASR corpus to get the most improvement from fine-tuning.

\section{Experiments}
\label{sec:exp}
In this section, the experimental settings and results are presented. All experiments were performed on a desktop computer running 64-bit Ubuntu 22.04 LTS. The computer was equipped with an AMD Ryzen 5900X CPU and 64GB of memory, and an Nvidia GeForce RTX 3090 GPU with 24GB of GPU memory. Due to the use of a pre-trained LM, the hyperparameters of the architecture were fixed except for the hidden-layer size (32) and the drop-out ratio (10\%) of the dense encoder, which are set manually without an extensive hyper-parameter search. The Adafactor optimizer is also used with its default parameters and 1e-5 learning-rate suggested for fine-tuning MiniLMv2~\cite{wang2020minilmv2}. 

The referenceless metric was trained using a large self-supervised corpus incorporating unique transcription hypotheses from OpenAI's Whisper ASR model~\cite{radford2022robust} for each audio sample available at CMU MOSEI and MOSEAS datasets~\cite{cmumosei, cmumoseas}. This corpus consisted of 134 hours of speech from YouTube videos spoken by 2,645 individuals. Almost half of the speeches in this corpus were in English, while the remaining duration was distributed uniformly among French, Spanish, Portuguese, and German languages. The proposed metric was self-supervised using unique pairs of speech transcripts, with one hypothesis in each pair known to have a higher quality than the other based on the compression level. After removing inconsistent pairs, there were 800,340 self-supervised parallel ASR hypothesis pairs in total, of which 20\% were reserved as a validation-set for early-stopping. An additional batch of data from the Common Voice training dataset was used to facilitate semi-supervised training. This batch included 148,393 unique hypotheses transcribed with top commercial ASR(s), and their WER was calculated to evaluate the transcription quality. The trained referenceless metric was blind-tested on multiple speech datasets, including Common Voice (English, French, Spanish)~\cite{ardila2019common}, Libri-Speech (English)~\cite{panayotov2015librispeech}, and a proprietary dataset containing 11 hours of meetings in-between non-native English speakers. The transcription hypotheses were obtained from top commercial ASR engines, namely AWS, AppTek, Azure, Deepgram, Google, and OpenAI's Whisper-Large, for each speech segment in them. 

\begin{table*}[ht]
\centering 
\caption{The self and semi-supervised NoRefER metric's performance on Common Voice \cite{ardila2019common} and Libri-Speech~\cite{panayotov2015librispeech} datasets, and a proprietary dataset, against the perplexity obtained from XLM-RoBERTa~\cite{conneau2019unsupervised}, regarding their correlations with WER ranks and scores.}
\label{tab:commonvoice}
\begin{tabular*}{\textwidth}{lccccccc}\hline
\multicolumn{1}{c}{\multirow{2}{*}{\textbf{Test Dataset - Language}}}     & \multirow{2}{*}{\textbf{Model}} & \multicolumn{3}{l}{\textbf{Correlation with WER ranking}} & \multicolumn{3}{l}{\textbf{Correlations with WER score itself}} \\ \cline{3-8} 
\multicolumn{1}{c}{}                                      &                                 & \textbf{Pearson}  & \textbf{Spearman}  & \textbf{Kendall} & \textbf{Pearson}    & \textbf{Spearman}    & \textbf{Kendall}   \\ \hline\hline
\multirow{2}{*}{Common Voice Test - English} & NoRefER-Self                         & \textbf{0.56}                  &  \textbf{0.48}                  & \textbf{0.55}                 &  0.42                   & 0.33                     & 0.24                   \\
                                                          & NoRefER-Semi                         & 0.53                  &  0.44                  & 0.52                 &  \textbf{0.48}                   & \textbf{0.53}                     & \textbf{0.40}                   \\
                                                          & XLMR-Large                            & 0.26                  &  0.22                  &  0.26                &  0.02                   & 0.21                     & 0.15                   \\ \hline
\multirow{2}{*}{Common Voice Test - French} & NoRefER-Self                         & 0.48                  &  \textbf{0.40}                  & 0.48                &   0.38                  & 0.33
                     & 0.24                   \\
                                                          & NoRefER-Semi                         & \textbf{0.48}                  &  0.39                  & \textbf{0.48}                 &  \textbf{0.50}                   & \textbf{0.56}                     & \textbf{0.41}                   \\
                                                          & XLMR-Large                            &  0.20                 & 0.17                   & 0.20                 & 0.02                    & 0.20                     &  0.14                  \\ \hline
\multirow{2}{*}{Common Voice Test - Spanish} & NoRefER-Self                         & \textbf{0.58}                  & \textbf{0.52}                   & \textbf{0.58}                  &  0.49                   &  0.40                    & 0.30                   \\
                                                          & NoRefER-Semi                         & 0.56                  &  0.48                  & 0.55                 &  \textbf{0.53}                   & \textbf{0.45}                     & \textbf{0.34}                   \\
                                                          & XLMR-Large                            &  0.25                 & 0.22                   & 0.25                 &  -0.01                    & 0.20                     &  0.14                  \\ \hline
\multirow{2}{*}{Libri-Speech Test Clean - Eng.}     & NoRefER-Self                         & 0.42                  & 0.35                   & 0.42                 &  0.30                   & 0.13                     &  0.09                  \\
                                                          & NoRefER-Semi                         & \textbf{0.57}                  &  \textbf{0.47}                  & \textbf{0.56}                 &  \textbf{0.42}                   & \textbf{0.51}                     & \textbf{0.37}                   \\
                                                          & XLMR-Large                            &    0.22                & 0.17                   &  0.21                & -0.06                    & 0.13                     &   0.09                \\ \hline
\multirow{2}{*}{Proprietary - Non-Native Eng.}     & NoRefER-Self                         & 0.37                  &    0.31                &  0.37                &   0.10                  &    0.33                  &  0.22                  \\
                                                          & NoRefER-Semi                         & \textbf{0.38}                  & \textbf{0.32}                  & \textbf{0.38}                 &  \textbf{0.16}                 &  \textbf{0.44}                    &        \textbf{0.31}            \\
                                                          & XLMR-Large                            &    0.29                &  0.25                  & 0.30                 &  -0.02                  &        -0.02              &   -0.01                \\ \hline
\end{tabular*}
\end{table*}

As a baseline, the referenceless metric was compared with the perplexity metric obtained from the state-of-the-art multilingual LM, XLM-RoBERTa Large~\cite{conneau2019unsupervised}. The perplexity metric measures the average number of predictions a language model must make to generate a speech transcript. A lower perplexity indicates a higher quality speech transcript, suggesting that the language model is more confident in its predictions. Table~\ref{tab:commonvoice} compares the proposed referenceless metric with the perplexity metric on different datasets and languages; where it consistently outperformed the baseline in all the blind test datasets, demonstrating its superiority in estimating ASR output quality. The table presents various correlation coefficients obtained by the proposed referenceless metric with the actual WER ranks and scores in datasets from multiple languages. The Pearson correlation coefficient measures the linear relationship between two variables, the Spearman correlation coefficient is a non-parametric measure of the monotonic relationship between two variables, and the Kendall correlation coefficient measures the agreement between two rankings and is non-parametric. For all correlation coefficients, the scores range from -1 to 1, with -1 indicating a strong negative correlation, 0 indicating no correlation, and 1 indicating a strong positive correlation~\cite{dancey2007statistics}. The default SciPy \cite{virtanen2020scipy} implementations is used for their calculation.

In the ablation study, the performance of NoRefER was compared using two different training methods. Firstly, NoRefER was trained using only self-supervised learning on a referenceless dataset of pairs, as described in Section~\ref{sec:datasetprep} (NoRefER-Self). Secondly, NoRefER was trained using the multi-tasked semi-supervised learning approach proposed (NoRefER-Semi). As shown in Table~\ref{tab:commonvoice}, the study results indicated a strong correlation between the two techniques. However, the NoRefER-Self model had limitations regarding inter-sample comparisons, as it showed weaker correlations with WER scores. In contrast, the NoRefER-Semi model had less correlation with WER ranking and was dependent on supervised data. The self-supervised NoRefER model was fine-tuned in 26 minutes and 47 seconds, while the semi-supervised version took 59 minutes and 16 seconds. The experiments measured an average duration of 0.095 seconds model inference time per hypothesis sample, which was the same for the both Self and Semi versions of NoRefER.

The pre-trained NoRefER weights and architecture, and the Python scripts required for reproducing all experimental results are open-sourced at this  repository: \url{https://anonymous.4open.science/r/Interspeech-NoRefER-6887}

\section{Discussion}
The experimental results demonstrate the effectiveness of the proposed referenceless quality metric and its potential impact in the ASR field. The evaluation of the referenceless metric revealed a strong correlation with the uncased and unpunctuated WER ranks and scores of the top commercial engines across samples. The high correlation with WER scores and ranks indicates the inter-sample and intra-sample reliability of the referenceless metric, respectively. When the contrastive learning is only performed between intra-sample hypotheses via self-supervision, NoRefER-Self achieves lower correlations with WER ranks than WER scores. In contrast, NoRefER-Semi also correlates well with WER scores due to the supervised inter-sample pairs used in its training. The experimental findings demonstrate that the referenceless metric can provide meaningful quality comparisons between different ASR models, serving as a viable alternative evaluation metric for ASR systems. Moreover, these results confirm the applicability of the referenceless metric over reference-based ones when comparing or A/B testing multiple ASR models/versions on production samples (for which ground-truth references are often not available) and selecting challenging ones for debugging and post-editing.

\section{Conclusion}
This work proposes a multi-lingual referenceless quality metric for ASR systems, which is useful when no ground truth is available to evaluate the quality of ASR outputs. The approach uses a pre-trained language model (MiniLMv2), fine-tuned on two datasets of varying quality using contrastive learning and pair-wise ranking. The results of the experiments demonstrate that the referenceless metric outperforms traditional reference-based and perplexity metrics from pre-trained language models. In addition, when tested on a blind dataset containing outputs from commercial ASR engines, the referenceless metric exhibits a strong correlation with the WER scores and ranks of these engines. This referenceless metric has the potential to significantly enhance the improvement and evaluation lifecycle of ASR systems in real-world applications. The future-research can focus on transfer-learning the proposed metric by incorporating audio-based features for referenceless quality estimation.

\bibliographystyle{IEEEtran}
\bibliography{mybib}

\begin{thebibliography}{10}
\providecommand{\url}[1]{#1}
\csname url@samestyle\endcsname
\providecommand{\newblock}{\relax}
\providecommand{\bibinfo}[2]{#2}
\providecommand{\BIBentrySTDinterwordspacing}{\spaceskip=0pt\relax}
\providecommand{\BIBentryALTinterwordstretchfactor}{4}
\providecommand{\BIBentryALTinterwordspacing}{\spaceskip=\fontdimen2\font plus
\BIBentryALTinterwordstretchfactor\fontdimen3\font minus
  \fontdimen4\font\relax}
\providecommand{\BIBforeignlanguage}[2]{{%
\expandafter\ifx\csname l@#1\endcsname\relax
\typeout{** WARNING: IEEEtran.bst: No hyphenation pattern has been}%
\typeout{** loaded for the language `#1'. Using the pattern for}%
\typeout{** the default language instead.}%
\else
\language=\csname l@#1\endcsname
\fi
#2}}
\providecommand{\BIBdecl}{\relax}
\BIBdecl

\bibitem{malik2021automatic}
M.~Malik, M.~K. Malik, K.~Mehmood, and I.~Makhdoom, ``Automatic speech
  recognition: a survey,'' \emph{Multimedia Tools and Applications}, vol.~80,
  pp. 9411--9457, 2021.

\bibitem{aldarmaki2022unsupervised}
H.~Aldarmaki, A.~Ullah, S.~Ram, and N.~Zaki, ``Unsupervised automatic speech
  recognition: A review,'' \emph{Speech Communication}, 2022.

\bibitem{fan2019neural}
K.~Fan, J.~Wang, B.~Li, S.~Zhang, B.~Chen, N.~Ge, and Z.~Yan, ``Neural
  zero-inflated quality estimation model for automatic speech recognition
  system,'' \emph{Proceedings of the International Speech Communication
  Association Conference, INTERSPEECH}, pp. 606--610, 2019.

\bibitem{radford2022robust}
A.~Radford, J.~W. Kim, T.~Xu, G.~Brockman, C.~McLeavey, and I.~Sutskever,
  ``Robust speech recognition via large-scale weak supervision,'' \emph{arXiv
  preprint arXiv:2212.04356}, 2022.

\bibitem{Chen_2021_CVPR}
X.~Chen and K.~He, ``Exploring simple siamese representation learning,'' in
  \emph{Proceedings of the IEEE/CVF Conference on Computer Vision and Pattern
  Recognition (CVPR)}, June 2021, pp. 15\,750--15\,758.

\bibitem{conneau2019unsupervised}
A.~Conneau, K.~Khandelwal, N.~Goyal, V.~Chaudhary, G.~Wenzek, F.~Guzm{\'a}n,
  E.~Grave, M.~Ott, L.~Zettlemoyer, and V.~Stoyanov, ``Unsupervised
  cross-lingual representation learning at scale,'' \emph{arXiv preprint
  arXiv:1911.02116}, 2019.

\bibitem{rabiner1993fundamentals}
L.~Rabiner and B.-H. Juang, \emph{Fundamentals of speech recognition}.\hskip
  1em plus 0.5em minus 0.4em\relax Prentice-Hall, Inc., 1993.

\bibitem{kalgaonkar2015estimating}
K.~Kalgaonkar, C.~Liu, Y.~Gong, and K.~Yao, ``Estimating confidence scores on
  asr results using recurrent neural networks,'' in \emph{2015 IEEE
  International Conference on Acoustics, Speech and Signal Processing
  (ICASSP)}.\hskip 1em plus 0.5em minus 0.4em\relax IEEE, 2015, pp. 4999--5003.

\bibitem{swarup2019improving}
P.~Swarup, R.~Maas, S.~Garimella, S.~H. Mallidi, and B.~Hoffmeister,
  ``Improving asr confidence scores for alexa using acoustic and hypothesis
  embeddings,'' \emph{Proceedings of the International Speech Communication
  Association Conference, INTERSPEECH}, 2019.

\bibitem{qiu2021learning}
D.~Qiu, Q.~Li, Y.~He, Y.~Zhang, B.~Li, L.~Cao, R.~Prabhavalkar, D.~Bhatia,
  W.~Li, K.~Hu \emph{et~al.}, ``Learning word-level confidence for subword
  end-to-end asr,'' in \emph{ICASSP 2021-2021 IEEE International Conference on
  Acoustics, Speech and Signal Processing (ICASSP)}.\hskip 1em plus 0.5em minus
  0.4em\relax IEEE, 2021, pp. 6393--6397.

\bibitem{del2018speaker}
M.~A. Del-Agua, A.~Gimenez, A.~Sanchis, J.~Civera, and A.~Juan,
  ``Speaker-adapted confidence measures for asr using deep bidirectional
  recurrent neural networks,'' \emph{IEEE/ACM Transactions on Audio, Speech,
  and Language Processing}, vol.~26, no.~7, pp. 1198--1206, 2018.

\bibitem{jiang2005confidence}
H.~Jiang, ``Confidence measures for speech recognition: A survey,''
  \emph{Speech communication}, vol.~45, no.~4, pp. 455--470, 2005.

\bibitem{ali2020word}
A.~Ali and S.~Renals, ``Word error rate estimation without asr output:
  E-wer2,'' \emph{Proceedings of the International Speech Communication
  Association Conference, INTERSPEECH}, pp. 616--620, 2020.

\bibitem{sheshadri2021bert}
A.~K. Sheshadri, A.~R. Vijjini, and S.~Kharbanda, ``Wer-bert: Automatic wer
  estimation with bert in a balanced ordinal classification paradigm,''
  \emph{Proceedings of European Chapter of the Association for Computational
  Linguistics Conference, EACL}, pp. 606--610, 2019.

\bibitem{namazifar2021correcting}
M.~Namazifar, J.~Malik, L.~E. Li, G.~Tur, and D.~H. T{\"u}r, ``Correcting
  automated and manual speech transcription errors using warped language
  models,'' \emph{Proceedings of the International Speech Communication
  Association Conference, INTERSPEECH}, pp. 921--925, 2021.

\bibitem{mathur2020results}
N.~Mathur, J.~Wei, M.~Freitag, Q.~Ma, and O.~Bojar, ``Results of the wmt20
  metrics shared task,'' in \emph{Proceedings of the Fifth Conference on
  Machine Translation}, 2020, pp. 688--725.

\bibitem{rei-etal-2020-unbabels}
R.~Rei, C.~Stewart, A.~C. Farinha, and A.~Lavie, ``Unbabel{'}s participation in
  the {WMT}20 metrics shared task,'' in \emph{Proceedings of the Fifth
  Conference on Machine Translation}, Nov. 2020, pp. 911--920.

\bibitem{wang2020minilmv2}
W.~Wang, H.~Bao, S.~Huang, L.~Dong, and F.~Wei, ``Minilmv2: Multi-head
  self-attention relation distillation for compressing pretrained
  transformers,'' \emph{arXiv preprint arXiv:2012.15828}, 2020.

\bibitem{shazeer2018adafactor}
N.~Shazeer and M.~Stern, ``Adafactor: Adaptive learning rates with sublinear
  memory cost,'' in \emph{International Conference on Machine Learning}.\hskip
  1em plus 0.5em minus 0.4em\relax PMLR, 2018, pp. 4596--4604.

\bibitem{cmumosei}
A.~B. Zadeh, P.~P. Liang, S.~Poria, E.~Cambria, and L.-P. Morency, ``Multimodal
  language analysis in the wild: Cmu-mosei dataset and interpretable dynamic
  fusion graph,'' in \emph{Proceedings of the 56th Annual Meeting of the
  Association for Computational Linguistics (Volume 1: Long Papers)}, 2018, pp.
  2236--2246.

\bibitem{cmumoseas}
A.~Zadeh, Y.~S. Cao, S.~Hessner, P.~P. Liang, S.~Poria, and L.-P. Morency,
  ``Cmu-moseas: A multimodal language dataset for spanish, portuguese, german
  and french,'' in \emph{Proceedings of the Conference on Empirical Methods in
  Natural Language Processing. Conference on Empirical Methods in Natural
  Language Processing}, vol. 2020.\hskip 1em plus 0.5em minus 0.4em\relax NIH
  Public Access, 2020, p. 1801.

\bibitem{ardila2019common}
R.~Ardila, M.~Branson, K.~Davis, M.~Henretty, M.~Kohler, J.~Meyer, R.~Morais,
  L.~Saunders, F.~M. Tyers, and G.~Weber, ``Common voice: A
  massively-multilingual speech corpus,'' \emph{arXiv preprint
  arXiv:1912.06670}, 2019.

\bibitem{panayotov2015librispeech}
V.~Panayotov, G.~Chen, D.~Povey, and S.~Khudanpur, ``Librispeech: an asr corpus
  based on public domain audio books,'' in \emph{2015 IEEE international
  conference on acoustics, speech and signal processing (ICASSP)}.\hskip 1em
  plus 0.5em minus 0.4em\relax IEEE, 2015, pp. 5206--5210.

\bibitem{dancey2007statistics}
C.~P. Dancey and J.~Reidy, \emph{Statistics without maths for
  psychology}.\hskip 1em plus 0.5em minus 0.4em\relax Pearson education, 2007.

\bibitem{virtanen2020scipy}
P.~Virtanen, R.~Gommers, T.~E. Oliphant, M.~Haberland, T.~Reddy, D.~Cournapeau,
  E.~Burovski, P.~Peterson, W.~Weckesser, J.~Bright \emph{et~al.}, ``Scipy 1.0:
  fundamental algorithms for scientific computing in python,'' \emph{Nature
  methods}, vol.~17, no.~3, pp. 261--272, 2020.

\end{thebibliography}

\end{document}